\documentclass[10pt,aps,pre,onecolumn,superscriptaddress,nofootinbib,longbibliography]{revtex4-2}

\usepackage[utf8]{inputenc}
\usepackage{amsmath,amssymb}
\usepackage{graphicx}
\usepackage{subcaption}
\usepackage[colorlinks=true,linkcolor=blue,citecolor=blue,urlcolor=blue]{hyperref}
\usepackage{comment}
\usepackage{titlesec}
\usepackage{indentfirst}

\usepackage{float}
\usepackage[english]{babel}
\usepackage[ruled,vlined]{algorithm2e}
\usepackage{booktabs}
\DontPrintSemicolon
\titleformat{\section}
  {\normalfont\Large\bfseries}{\thesection}{1em}{}

\titleformat{\subsection}
  {\normalfont\large\bfseries}{\thesubsection}{1em}{}
\titleformat{\subsubsection}
  {\normalfont\large\bfseries}{\thesubsubsection}{1em}{}
\usepackage[justification=raggedright,singlelinecheck=false]{caption}

\begin{document}

\title{Neural network optimization strategies and the topography of the loss landscape}

\author{Jianneng Yu}
\affiliation{Department of Physics \& Astronomy, Rutgers, The State University of New Jersey, 136 Frelinghuysen Rd., Piscataway, NJ 08854, U.S.A.}

\author{Alexandre V. Morozov}
\email{morozov@physics.rutgers.edu}
\affiliation{Department of Physics \& Astronomy, Rutgers, The State University of New Jersey, 136 Frelinghuysen Rd., Piscataway, NJ 08854, U.S.A.}

\begin{abstract}
Neural networks are trained by optimizing multi-dimensional sets of fitting parameters on non-convex loss landscapes. Low-loss regions of the landscapes correspond to the parameter sets that perform well on the training data. A key issue in machine learning is the performance of trained neural networks on previously unseen test data. Here, we investigate neural network training by stochastic gradient
descent (SGD) - a non-convex global optimization algorithm which relies only on the gradient of the objective function.
We contrast SGD solutions with those obtained via a non-stochastic quasi-Newton method, which utilizes curvature information to determine step direction and Golden Section Search to choose step size.
We use several computational tools to investigate neural network parameters obtained by these two optimization methods,
including kernel Principal Component Analysis and a novel, general-purpose algorithm for finding low-height paths between pairs of points on loss or energy landscapes, FourierPathFinder. We find that the choice of the optimizer profoundly affects the nature of the resulting solutions. SGD solutions tend to be separated by lower barriers than quasi-Newton solutions, even if both sets of solutions are regularized by early stopping to ensure adequate performance on test data. When allowed to fit extensively on the training data, quasi-Newton solutions occupy deeper minima on the loss landscapes that are not reached by SGD. These solutions are less generalizable to the test data however. Overall, SGD explores smooth basins of attraction, while quasi-Newton optimization is capable of finding deeper, more isolated minima
that are more spread out in the parameter space. Our findings help understand both the topography of the loss landscapes and the fundamental role of landscape exploration strategies in creating robust, transferrable neural network models.
\end{abstract}

\maketitle

\section{Introduction}

The problem of finding maximum or minimum values of multi-dimensional functions with complex non-linear structure arises in engineering, economic and financial forecasting, biological data analysis, molecular physics, robot design, and numerous other scientific and technological settings.
Notable examples include finding free energy minima in computer simulations of protein folding~\cite{Onuchic:2004,Dill:2008}, converging to high-fitness states in evolving populations~\cite{Crow1970,Kimura1983,Gillespie2004},
and minimizing loss functions in training neural networks~\cite{Goodfellow2016,Mehta:2018dln}.

Modern optimization algorithms used in deep learning~\cite{Nocedal_Wright_2006,ruder2017} are powerful enough to drive neural network (NN) training loss to very low values across a wide variety of NN architectures and training datasets. However, minimizing training error does not necessarily guarantee robust performance on unseen data~\cite{10.1145/3446776}. One reason for this is the possibility of overfitting in overparameterized deep NNs, which is typically mitigated by early stopping, weight regularization, and other techniques~\cite{Goodfellow2016,Mehta:2018dln}. Some sets of trained NN parameters generalize well, while others, despite achieving equally low training loss, fail to transfer, resulting in poor test accuracy~\cite{keskar2017largebatchtrainingdeeplearning}.
This behavior underscores a central challenge in deep learning: understanding why some models generalize while others do not, even if they occupy low-loss regions of the training landscape.

A key open question in deep learning theory is how different training procedures influence the types of solutions found and the relative positions of these solutions on the loss landscape; a related area of inquiry focuses on the loss landscape topography and connectivity~\cite{Chaudhari_2019,jastrzębski2018factorsinfluencingminimasgd,Feng:2021}. For highly non-linear systems such as neural networks, the loss surface is thought to be composed of numerous basins of attraction connected by relatively flat valleys~\cite{wei2019noiseaffectshessianspectrum,Feng:2021}.
Interestingly, even linear interpolation between trained models can reveal key features of the loss landscape geometry, including the topography of barriers and valleys between local minima~\cite{goodfellow2015qualitativelycharacterizingneuralnetwork}. 
Later work addressed loss landscape visualization with a variety of computational tools, establishing a dependence of the loss landscape curvature around optimized solutions on the training method used in the optimization~\cite{NEURIPS2018_a41b3bb3}.

It has been widely observed that stochastic gradient descent (SGD) with small batch sizes tends to converge to flatter minima, which are often associated with better generalization performance~\cite{keskar2017largebatchtrainingdeeplearning, wei2019noiseaffectshessianspectrum,Feng:2021}. In contrast, adaptive methods like Adam~\cite{kingma2014adam,Mehta:2018dln} and RMSProp~\cite{tieleman2012rmsprop,Mehta:2018dln}, though potentially faster in convergence and lower in training loss, frequently lead to sharper solutions that may generalize less effectively~\cite{10.5555/3294996.3295170}. Other methods that rely on curvature information can achieve even lower training loss, but are also known to converge to sharper minima~\cite{10.5555/3327345.3327403}. These observations prompt fundamental questions: do approaches that rely solely on the gradient information, such as SGD, find solutions in the same broad region of parameter space, or do they produce qualitatively distinct minima with varying generalization properties? Are there significant differences between solutions found by gradient vs. quasi-Newton methods~\cite{Bishop_2006,Nocedal_Wright_2006} which rely on the curvature information? Addressing these questions is crucial for understanding how optimization algorithms explore high-dimensional, non-convex loss landscapes which have to be traversed in NN training.

Several studies have recently examined the structure of NN loss
landscapes and diversity, connectivity, and generalizability of optimized solutions. For instance, the distribution of Hessian eigenvalues during training was found to have significant variation across optimization strategies~\cite{pmlr-v97-ghorbani19b}, suggesting that different methods may settle into different types of low-loss regions. It was argued that the minima of the loss function are connected by low-height paths~\cite{10.5555/3327546.3327556,pmlr-v80-draxler18a}, consistent with high levels of connectivity and continuity on the loss landscapes.
Furthermore, it was demonstrated that loss landscapes in deep neural networks admit star-convex paths between initial states and optimized solutions, allowing gradient-based methods such as SGD to avoid local kinetic traps~\cite{zhou2019sgdconvergesglobalminimum}. Sets of optimized solutions in overparameterized networks frequently form star domains, which are regions where any point can be connected to a `central' solution via low-loss paths~\cite{sonthalia2024deepneuralnetworksolutions}.


In this work, we focus on the relative advantages and disadvantages of employing SGD, a widely used stochastic gradient-based optimizer~\cite{Mehta:2018dln}, versus an efficient quasi-Newton method,
Limited-memory Broyden-Fletcher-Goldfarb-Shanno~\cite{Broyden1970,Fletcher1970,Goldfarb1970,Shanno1970,Nocedal_Wright_2006},
augmented with a Golden Section Search for determining the step size (L-BFGS-GSS). We find that
although both methods yield similar performance on test data when early stopping is employed for regularization, the solutions found by the two optimizers are
qualitatively different. The L-BFGS-GSS solutions are separated by higher barriers and are more distant from one another in parameter space. L-BFGS-GSS optimization can lead to very low training loss values compared to SGD, resulting in overfit, poorly generalizable solutions. In contrast, SGD solutions are more generalizable, in agreement with previous studies~\cite{keskar2017largebatchtrainingdeeplearning, wei2019noiseaffectshessianspectrum,Feng:2021}. To study SGD and L-BFGS-GSS solutions sets, we develop
a number of approaches aimed at their visualization in the context of the loss landscape topography, including a novel algorithm for finding low-height paths connecting pairs of points on loss, energy, or negative fitness landscapes, \texttt{FourierPathFinder}.



\section{Methods}

\subsection{Neural Network Architectures}

We study the loss landscapes of four neural network (NN) architectures trained on the MNIST dataset of $28 \times 28$ black-and-white images of handwritten digits~\cite{6296535}. The first NN we consider is a fully connected perceptron (FCP)
-- a feedforward network with two hidden layers of 50 units each, which use ReLU activations and no bias terms. The input layer has 784 nodes and receives $D = 28 \times 28 = 784$ pixel values as inputs; the output layer is a \texttt{softmax} classifier into 10 single-digit classes: $0 \dots 9$.
The second NN is a convolutional neural network (CNN)~\cite{NIPS1988_a97da629} with LeNet architecture~\cite{726791}.
It consists of two convolutional layers with 6 and 16 channels, respectively, each followed by average pooling and ReLU activation. The two-layer convolutional block is followed by three fully connected layers with 120, 84, and 10 nodes; the last layer outputs probabilities of 10 single-digit classes.

The third NN is a Long Short-Term Memory (LSTM) recurrent network~\cite{10.1162/neco.1997.9.8.1735}. Following Ref.~\cite{le2015simplewayinitializerecurrent}, each image is flattened and permuted over all 784 pixels (with a fixed random seed for reproducibility), then reshaped into 28 time steps of 28 features each for the sequential processing by the LSTM. The LSTM has a hidden layer size of 48 and outputs the final hidden state to a linear classifier with 10 output units and no bias term.
Finally, we employ a shallow autoencoder architecture~\cite{10.5555/104279.104293} to explore unsupervised loss landscapes. The encoder and decoder each consist of two FC layers with 32 units per layer, \texttt{softplus} activations, and no biases; the final layer of the decoder uses a sigmoid activation to reconstruct the image.
Implementations of all four NN model architectures are available at \texttt{https://github.com/jy856-jpg/path-finding}. The total number of fitting parameters for each NN, $N_\text{prm}$, is listed in Table~\ref{loss:table}.

For the three classification models, we define the loss function $\langle l(x^\text{train},\omega) \rangle$ or $\langle l(x^\text{test},\omega) \rangle$ as the cross-entropy loss between the predicted output and the true label, averaged over the entire training or test set (except for SGD optimization, where the averages are taken over 64 images in a mini-batch). For the autoencoder, $\langle l(x^\text{train},\omega) \rangle$ or $\langle l(x^\text{test},\omega) \rangle$ is defined as the mean squared reconstruction error, also averaged over the entire training or test set. Here, $\omega$ denotes a vector of model-dependent weights and biases. We use a standard split of the MNIST dataset into $\{ x^\text{train} \}$ with $N=5 \times 10^4$ training images and $\{ x^\text{test} \}$ with $N=10^4$ test images~\cite{6296535}.



\subsection{Quasi-Newton optimization with Golden Section Search}

Second-order optimization methods aim to improve convergence by incorporating curvature information~\cite{Bishop_2006,press2007numerical}. The function to be minimized is approximated locally using a second-order Taylor expansion:
\begin{equation}
    f(z+\epsilon) \approx f(z)+\nabla f(z)^T \epsilon +\frac{1}{2} \epsilon^T H \epsilon,
    \label{taylor_exp}
\end{equation}
where $f(z)$ is a function of an $\mathcal{N}$-dimensional argument $z$, $\nabla f(z) = \nabla_y f(y) \vert_{y=z}$ represents its gradient evaluated at $z$
($\nabla_y = \partial/\partial y_1 \dots \partial/\partial y_\mathcal{N}$),
and $H$ is the Hessian matrix of second derivatives:
$H_{ij} = \partial^2 f(y)/\partial y_i \partial y_j \vert_{y=z}$ ($i,j=1 \dots \mathcal{N}$).

In the Newton-Raphson iterative optimization method, the update step is given by~\cite{Bishop_2006}:
\begin{equation}
    z^\text{new} = z^\text{old} - H^{-1} \nabla f(z).
    \label{Inverse Hess}
\end{equation}

Note that both the direction and the magnitude of the minimization step are determined
by the inverse of the Hessian matrix -- there is no need to choose the step size as in first-order optimization methods that rely solely on the gradients~\cite{Mehta:2018dln}. 
However, computing and inverting the Hessian matrix scales poorly with the number of parameters~\cite{Bishop_2006,doi:10.1137/16M1080173}, which makes it infeasible to apply Eq.~\eqref{Inverse Hess} to large models such as deep neural networks. To mitigate the high computational cost of second-order methods, quasi-Newton algorithms such as Limited-memory Broyden-Fletcher-Goldfarb-Shanno (L-BFGS)~\cite{Broyden1970,Fletcher1970,Goldfarb1970,Shanno1970,Nocedal_Wright_2006} are used to approximate the product between the inverse Hessian and the gradient without explicitly computing and inverting the Hessian matrix.
After computing the search direction $p = H^{-1} \nabla f(z)$,
a line search is typically carried out to find the optimal step size,
using Armijo-Wolfe conditions or other suitable criteria~\cite{Nocedal_Wright_2006,doi:10.1137/1011036}.

In the NN context, the goal is to optimize $f(\omega) = \langle l(x, \omega) \rangle$,
where $l(x,\omega)$ denotes the loss function corresponding to the input datapoint $x$ (in our case, a single MNIST image) at the current
set of weights $\omega$, and the average is taken over all images in the training
dataset. Note that $\mathcal{N} = N_\text{prm}$ in this case (cf. Table~\ref{loss:table}).
Here, we use the Golden Section Search (GSS)~\cite{press2007numerical} -- 
a lightweight and derivative-free approach to line search. 
The goal of GSS, as in any line search algorithm, is to find an optimal step size $u$ along the search direction $p$ by minimizing $f(u) = \langle l(x, \omega-u p) \rangle$,
where $f(u)$ is now a 1D function. Briefly, GSS is used to find the minimum of a unimodal 1D function on a closed interval by repeatedly shrinking a bracket that contains the minimum. At each step, the interval is reduced by a fixed fraction determined by the golden ratio $\phi = {(1 + \sqrt{5})}/{2}$, allowing one function evaluation to be reused, so that only a single new evaluation is needed per iteration. The algorithm converges robustly and linearly, with
the error decreasing by a constant factor at each iteration.

We call the L-BFGS algorithm augmented by GSS the L-BFGS-GSS optimizer, summarized in Algorithm~\ref{algo:1}.
Note that our customized implementation of the L-BFGS optimizer maintains two lists of size $m$, $s_\text{list}$ and $y_\text{list}$, that contain $m$ most recent differences $s_k=\omega_k-\omega_{k-1}$ and
$y_k = \nabla f(\omega_{k}) - \nabla f(\omega_{k-1})$,
respectively.

\begin{algorithm}[H]
\caption{Overview of the L-BFGS-GSS optimizer.}
\KwIn { Objective function $f(\omega) = \langle l(x,\omega) \rangle$, initial NN parameters $\omega_0$, maximum number of iterations $K$, training data $x$ }
\KwOut { Optimized NN parameters $\omega^\star$ }
\For{$k = 1$ $\mathrm{to}$ $K$}{
     Compute gradient, $\nabla f(\omega_k)$  \;
     Compute search direction $p(\nabla f(\omega_k), s_{\mathrm{list}}, y_{\mathrm{list}})$ via L-BFGS  \;
     Determine optimal step size $u^\star$ with GSS, such that $f(\omega_k - u^\star p)$ is minimized  \;
     $\omega_{k+1} \gets \omega_k - u^\star p$ \;
     Update $s_{\mathrm{list}}$, $y_{\mathrm{list}}$ \;
 \textbf{return} $\omega^* \gets \omega_{K+1}$ \;
 }
\label{algo:1}
\end{algorithm}

\subsection{Construction of low-loss paths between two points on the landscape}


We have developed a general-purpose path-finding algorithm, called \texttt{FourierPathFinder} (Algorithm~\ref{algo:FPF}), which constructs low-loss paths between two points on a multi-dimensional landscape as a combination of a straight line and a truncated Fourier series:
\begin{equation}
    \omega(t)=t \omega^i + (1-t) \omega^j + \sum_{n=1}^{N_\text{F}} b_n \sin(n\pi t),
    \label{Path:combined}
\end{equation}
where $\omega^i$ and $\omega^j$ are the initial and final points on the landscape,
$t \in [0, 1]$ is the curve parameter, and $N_\text{F}$ is the total number of Fourier terms
(we typically set $N_\text{F}=10$). The Fourier coefficients $b_n$ are initialized to $0$.
For NN loss landscapes, we discretize the curve parameter $t$ into $M=50$ equally spaced values.






The total loss along the path in Eq.~\eqref{Path:combined} is computed as:
\begin{equation}
    L(\omega^i,\omega^j) = \sum^M_{m=1} \langle l(x,\omega(t_m)) \rangle + \lambda \sum^{M-1}_{m=1} |\omega(t_{m+1})-\omega(t_{m})|^2,
    \label{Total Loss}
\end{equation}
where the first term is the cumulative loss along the path. The second term is a regularization penalty, scaled by the hyperparameter $\lambda$ which controls the smoothness and the non-linearity of the path. We choose $\lambda = 10^{-4}$ for the paths on NN loss landscapes -- we find that this value provides a reasonable balance between
the total path length and the cumulative loss along the path. \\


\begin{algorithm}[H]
\caption{Overview of the \texttt{FourierPathFinder} algorithm.}
\KwIn{ Path loss function $L(\omega^i,\omega^j)$ (Eq.~\eqref{Total Loss}), maximum number of iterations $K$, input data $\{ x \}$, curve parameter values $\{ t_m \}_{m=1}^M$, regularization coefficient $\lambda$, Fourier coefficients $\{ b_n \}_{n=1}^{N_\text{F}}$ initialized to 0. }
\KwOut{ Optimized Fourier coefficients $\{ b_n^\star \}_{n=1}^{N_\text{F}}$. }
\For{$k = 1$ \KwTo $K$}{
    Compute NN loss gradients with respect to NN parameters at each $t_m$: $g_m(x)=\frac{\partial \langle l(x,\omega) \rangle}{\partial \omega} \vert_{\omega=\omega(t_m)}$ \;
    Compute the path loss gradient with respect to the Fourier coefficients: $\frac{\partial L}{\partial b_n}=\sum^M_{m=1} g_m(x)\sin(n\pi t_m) + 2 \lambda \sum^{M-1}_{m=1} [\omega(t_{m+1})-\omega(t_{m})] [\sin(n\pi t_{m+1})-\sin(n\pi t_{m})]$ \;
    Update $\{ b_n \}_{n=1}^{N_\text{F}}$ using Adam optimizer~\cite{kingma2014adam} on the entire dataset $\{ x \}$\;
\textbf{return} $\{ b_n^\star \}_{n=1}^{N_\text{F}} \gets \{ b_n \}_{n=1}^{N_\text{F}}$ \;
}
   \label{algo:FPF}
\end{algorithm}

To characterize barrier height along a given path, we compute path height,
defined as the maximum loss encountered along the discretized trajectory:
\begin{equation}
    H = \max_{t_m} \left\{ \langle l(x,\omega(t_m)) \rangle \right\}.
    \label{H}
\end{equation}






\subsection{Dimensionality reduction for characterizing optimized NN parameter sets}

To visualize the training or test sets of NN parameters optimized using either L-BFGS or SGD, we employ a dimensionality reduction technique called kernel Principal Component Analysis (kPCA)~\cite{Scholkopf1998,Bishop_2006}. Briefly, kPCA is a nonlinear generalization of standard PCA, a linear dimensionality reduction method designed to identify orthogonal directions (principal components) along which the data varies most~\cite{Bishop_2006}. kPCA extends this approach to capture nonlinear structures in the data by implicitly mapping input datapoints $x \in \mathbb{R}^D$ into a feature space $\phi(x) \in \mathbb{R}^M$ through nonlinear mapping $x \to \phi(x)$.

The similarity between two points $x$ and $x'$ in the feature space is expressed through a kernel function $k(x,x') = \phi(x)^{T} \phi(x')$, which computes the inner product between
their feature-space representations. Common kernel functions include the linear kernel $k(x,x')=x^{T} x'$, the degree $n$ polynomial kernel $k(x,x')=(x^T x' + C)^n$, and the radial basis function (RBF) kernel $k(x,x') = \exp{(-|x-x'|^2/2\sigma^2)}$, each defining a different notion of similarity between datapoints $x$ and $x'$. Each kernel corresponds to a potentially infinite-dimensional set of feature vectors. Note also that kernels often depend on hyperparameters such as $C$ in the polynomial or $\sigma$ in the RBF kernel. We set $2\sigma^2 = N_\text{prm}$ in visualizing optimized vectors of NN weights and biases.

As is typical in kernel-based methods, kPCA avoids constructing the feature vectors explicitly -- the dimensionality reduction is carried out using the kernel matrix $K \in \mathbb{R}^{N \times N}$, where $K_{ij} = k(x_i,x_j)$ and $N$ is the number of datapoints. Specifically, the kernel matrix
is centralized~\cite{Bishop_2006}:
$K \to \widetilde{K}$, where the centralized kernel corresponds to
the feature vectors with zero mean: $\widetilde{K}_{ij} = \widetilde{k} (x_i,x_j) = \widetilde{\phi}(x_i)^{T} \widetilde{\phi}(x_j)$, with $\sum_{n=1}^N \widetilde{\phi}(x_n) = 0$. Next, the eigenvalues and eigenvectors of the $N \times N$ centralized kernel matrix are found by solving the eigenvalue problem: $\widetilde{K} \alpha^{(k)} = \lambda^{(k)} \alpha^{(k)}$. Finally, the principal component projections are computed using
$\text{PC}_k(x) = \sum_{i=1}^N \alpha_i^{(k)} k(x, x_i)$, where $k$ labels the eigenvalues and $x$ is the input vector to be projected.

\section{Results}


\subsection{Overview of NN optimization and loss landscape visualization}

\begin{figure}[!htb]
  \centering
  \includegraphics[width=\textwidth]{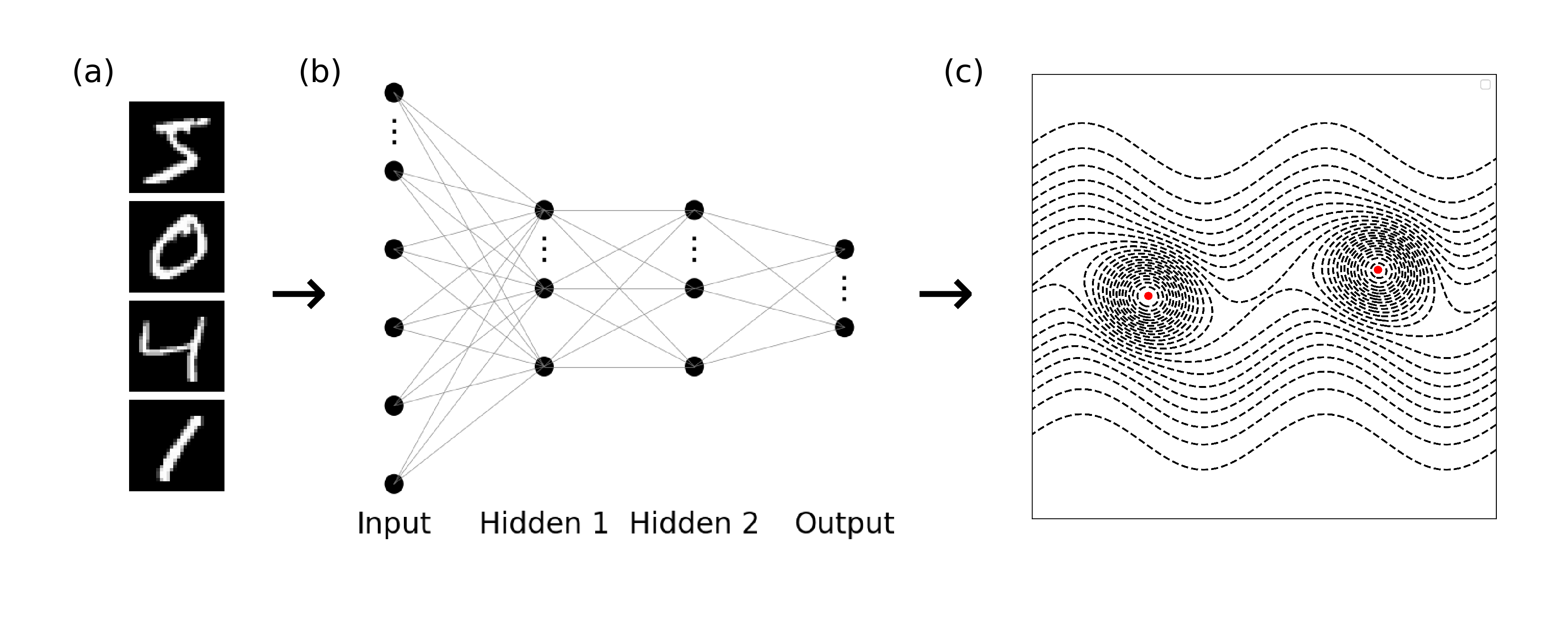}
  \caption{\textbf{Generation and visualization of optimized NN parameter sets.}
    (a) A subset of input training/test data. A single $28 \times 28$ MNIST image~\cite{6296535} is used as NN input.
    (b) Representative NN architecture, with an input layer, two hidden layers, and
    an output layer.
    (c) A conceptual sketch of the corresponding NN loss landscape, with two local minima (red dots) located in a shallow valley. The basins of attraction of the two minima are separated by a relatively low barrier.
    }
    \label{Fig:overview}
\end{figure}

Our approach to NN training and loss landscape exploration is outlined in Figure~\ref{Fig:overview}.
We train four NN architectures (FCP, LeNet CNN, Autoencoder, and LSTM) on a set of MNIST images~\cite{6296535} using two algorithms: Stochastic Gradient Descent (SGD)~\cite{doi:10.1137/16M1080173} with $64$ images per mini-batch and a customized quasi-Newton algorithm, L-BFGS-GSS (Methods). We obtain sets of optimized NN parameters located in low-loss regions of the multi-dimensional loss landscapes and study the depths of these minima, their basins of attraction, and the heights of the barriers separating optimized parameter vectors from one another. 

Specifically, we assemble four sets of optimized parameters for each of the four NN
architectures considered in this work: FCP, LeNet CNN, Autoencoder, and LSTM
(see Methods for NN implementation details). All neural nets employ $\mathcal{O} (10^4)$ fitting parameters (Table~\ref{loss:table}).
The first two sets, $\{ \omega^{\text{train},i}_{\text{BFGS}} \}_{i=1}^{48}$ and $\{ \omega^{\text{test},i}_{\text{BFGS}} \}_{i=1}^{48}$, correspond to the solutions obtained 
using the L-BFGS-GSS quasi-Newton algorithm without mini-batches (Algorithm~\ref{algo:1}; see Methods for details).
The second two sets, $\{ \omega^{\text{train},i}_{\text{SGD}} \}_{i=1}^{48}$ and $\{ \omega^{\text{test},i}_{\text{SGD}} \}_{i=1}^{48}$, comprise solutions obtained using SGD~\cite{Robbins_Monro_1951,ruder2017} with 64-image mini-batches. 
Each NN model is trained starting from 75 different random initializations of weights and biases for each optimizer type; 48 solutions with the lowest training loss from each method are selected to form $\{ \omega^{\text{train},i}_{\text{BFGS}} \}_{i=1}^{48}$ and $\{ \omega^{\text{train},i}_{\text{SGD}} \}_{i=1}^{48}$. 
To obtain the corresponding test set fitting weights, $\{ \omega^{\text{test},i}_{\text{BFGS}} \}_{i=1}^{48}$ and $\{ \omega^{\text{test},i}_{\text{SGD}} \}_{i=1}^{48}$, we employ early stopping at the minimum of test loss for the models included into our training sets~\cite{Mehta:2018dln}.

\begin{figure}[!htb]
  \centering
  \includegraphics[width=\textwidth]{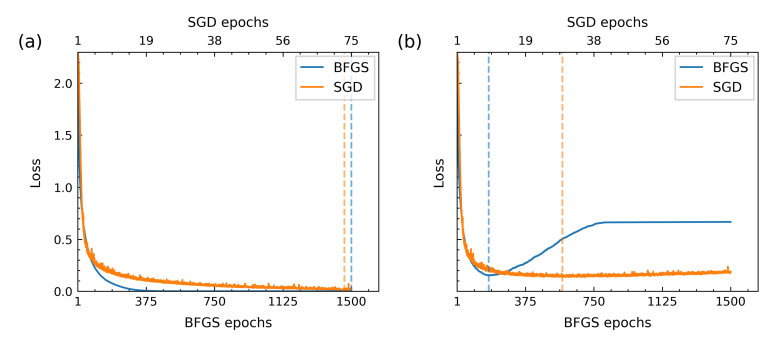}
  \caption{\textbf{LSTM loss curves}. Representative LSTM training (a) and test (b) loss curves ($\langle l(x^\text{train},\omega) \rangle$ and $\langle l(x^\text{test},\omega) \rangle$, respectively) as a function of the number of epochs. In both panels, dashed vertical lines mark the epochs where the loss curves of the same color reach their minima. The weight configurations $\omega^{\text{train}}_{\text{SGD}},\omega^{\text{test}}_{\text{SGD}}$ and $\omega^{\text{train}}_{\text{BFGS}},\omega^{\text{test}}_{\text{BFGS}}$ denote the sets of NN parameters found at these minima (optimized with SGD and L-BFGS-GSS, respectively). 
  }
  \label{loss:curves}
\end{figure}

Figure~\ref{loss:curves} shows a representative example of training/test L-BFGS-GSS and SGD loss curves for the LSTM network (loss curves for the other three architectures are displayed in Fig.~\ref{loss:curves:extra}). The training or test optimized NN parameter sets correspond to the minimum of the corresponding loss curve. Table~\ref{loss:table} shows the average loss over 48 independent runs. For all NN architectures, the values of the average training loss are considerably lower for the L-BFGS-GSS optimizer compared to SGD. However, the average test loss values are comparable, indicating that the L-BFGS-GSS training weights are likely to be overfit. Note also that the L-BFGS-GSS test loss curves rise sharply from the minima in all NN architectures except the autoencoder (cf. blue lines in Figs.~\ref{loss:curves}b and \ref{loss:curves:extra}b,d,f). In contrast, SGD weight sets appear to be more generalizable.

\begin{table}[h!]
\centering
\caption{\textbf{Average training and test loss.} For each NN architecture, we list the
training loss $\langle l(x^{\text{train}},\omega^{\text{train}}) \rangle$ and the test loss $\langle l(x^{\text{test}},\omega^{\text{test}}) \rangle$ averaged over $48$ independent runs. Also listed are $N_{\text{prm}}$, the total number of NN fitting parameters (weights and biases) in each of the four architectures.
}

\begin{tabular}{lccccc}
\toprule
& \multicolumn{2}{c}{$\langle \overline{l(x^{\text{train}},\omega^{\text{train}})} \rangle$} & \multicolumn{2}{c}{$\langle \overline{l(x^{\text{test}},\omega^{\text{test}})} \rangle$} \\
\cmidrule(lr){2-3} \cmidrule(lr){4-5}
\textbf{NN} & \textbf{L-BFGS} & \textbf{SGD} & \textbf{L-BFGS} & \textbf{SGD} & $N_{\text{prm}}$ \\\\
\midrule
FCP               & 1.26e-08 & 6.21e-04 & 9.69e-02 & 8.25e-02 & 42200 \\
LeNet             & 3.35e-08 & 1.76e-05 & 4.17e-02 & 3.89e-02 & 61706 \\
Autoencoder       & 1.69e-02 & 4.57e-02 & 1.65e-02 & 4.52e-02 & 52224 \\
LSTM              & 1.63e-06 & 1.35e-02 & 1.55e-01 & 1.28e-01 & 15456 \\
\bottomrule
\end{tabular}
\label{loss:table}
\end{table}

\subsection{Low-loss paths connecting optimized states on NN loss landscapes}

\begin{figure}[!htb]
\centering
\includegraphics[width=\textwidth]{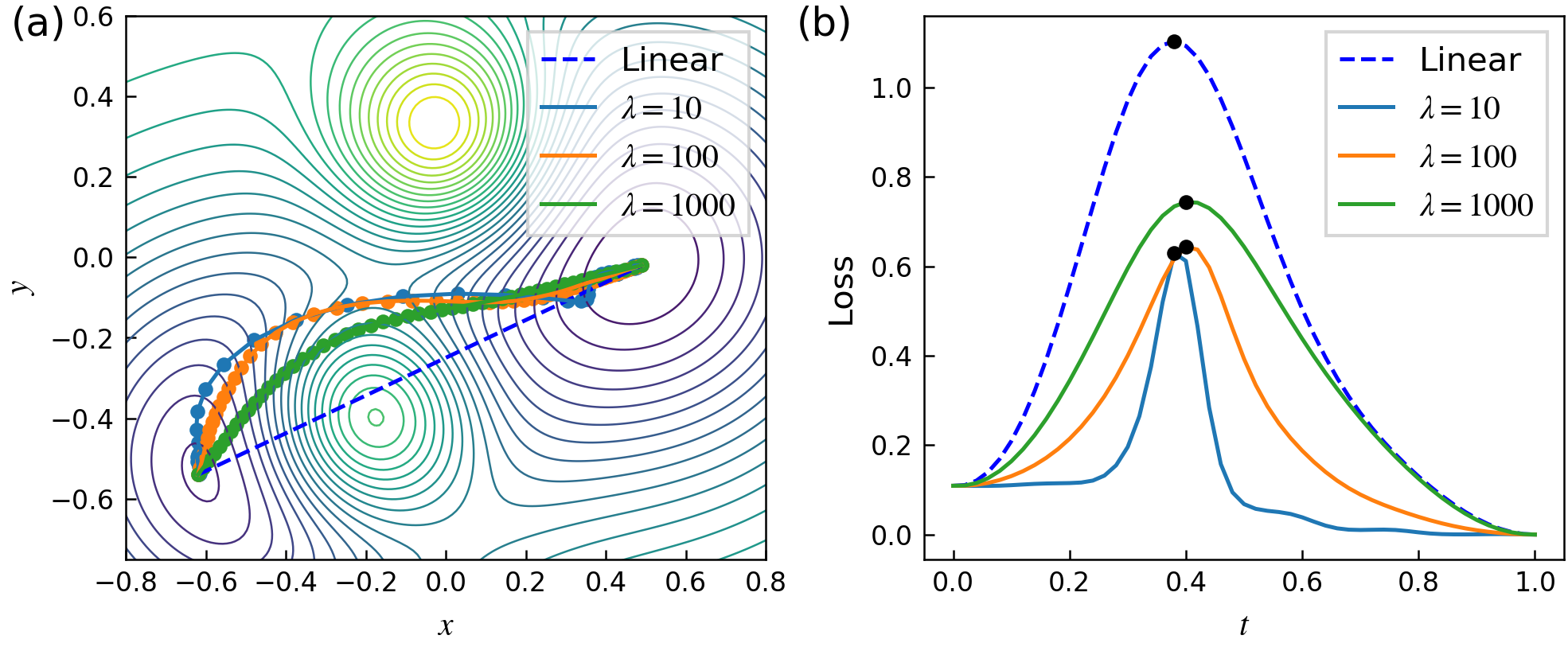}
\caption{\textbf{Low-loss paths on a 2D landscape.}
(a) Two-dimensional loss landscape composed of two positive and two negative Gaussian peaks:
$f(x, y) = -\sum_{i=1}^{2} \exp[-3|\mathbf{r} - \mathbf{c}_i|^2]
+ \sum_{j=1}^{2} \exp[-15|\mathbf{r} - \mathbf{d}_j|^2] + C$,
where $\mathbf{r} = (x,y)$, $\mathbf{c}_1 = (-0.5, -0.5)$, $\mathbf{c}_2 = (0.5, 0.0)$,
$\mathbf{d}_1 = (-0.2, -0.4)$, $\mathbf{d}_2 = (0.0, 0.3)$, and $C = 1.019$.
Four representative paths connecting two landscape minima: $\mathbf{w}_1 = (-0.62, -0.54)$ and $\mathbf{w}_2 = (0.49, -0.02)$ are shown: a linear interpolation path (dashed blue line) and three \texttt{FourierPathFinder} optimized paths ($\lambda = 10$, solid blue curve; $\lambda = 100$, solid orange curve; $\lambda = 1000$, solid green curve). Dots indicate function values at discrete time steps $t_m \in [0,1]$ along the path: $f(x(t_m),y(t_m))$, $m = 1 \dots M$ ($M = 100$).
(b) Loss values $f(x(t),y(t))$ as a function
of the curve parameter $t$ along the four paths in panel (a):
the linear interpolation path (dashed blue curve) and three \texttt{FourierPathFinder} paths (solid curves with the colors matching the paths in panel (a)).
Path heights $H_i$ (Eq.~\eqref{H}) are labeled with black dots, with
$H_0 = 1.102$ (straight line), $H_1 = 0.646$ (optimized path, $\lambda = 10$), $H_2 = 0.640$ (optimized path, $\lambda = 100$), and $H_3 = 0.652$ (optimized path, $\lambda = 1000$).
}
\label{PathLoss:synthetic}
\end{figure}

We have developed \texttt{FourierPathFinder}, an algorithm for finding low-height paths connecting two points on multi-dimensional loss or energy landscapes (Algorithm~\ref{algo:FPF}; see Methods for implementation details). Figure~\ref{PathLoss:synthetic}a illustrates our algorithm on a synthetic 2D landscape $f(x,y)$ with two local minima -- shown are a linear path between the two minima and three paths found using \texttt{FourierPathFinder} with increasing regularization penalties:
$\lambda = 10,100,1000$. Figure~\ref{PathLoss:synthetic}b traces the corresponding loss values along these paths, given here by the values of $f(x,y)$ along each parametrized curve.

We observe that all three optimized paths find a low-loss valley between two neighboring maxima. However, the length of the $\lambda = 10$ path is not sufficiently constrained, enabling it to spend more time in the low-loss regions around the two minima and make larger steps in crossing the barrier between the two basins of attraction (cf. the spacing of the blue points in Fig.~\ref{PathLoss:synthetic}a and the loss profile for the $\lambda = 10$ curve in Fig.~\ref{PathLoss:synthetic}b). Larger values of $\lambda$ prevent this non-uniform behavior and result in more regular paths. Despite these differences, the maximum height along the path (Eq.~\eqref{H}) is fairly
insensitive to $\lambda$. Thus, there is no need to fine-tune this hyperparameter.

Next, we consider the heights of the paths connecting optimized vectors of parameters on the NN loss landscapes. For each of the four sets containing $48$
vectors of trained parameters, we randomly choose $300$ pairs of vectors and use \texttt{FourierPathFinder} (with $\lambda = 10^{-4}$) to find the low-loss paths connecting them. We record the corresponding path heights (Eq.~\eqref{H}), which characterize the connectivity of optimized vectors $\omega$ in the parameter space.

\begin{figure}[!htb]
  \centering
  \includegraphics[width=\textwidth]{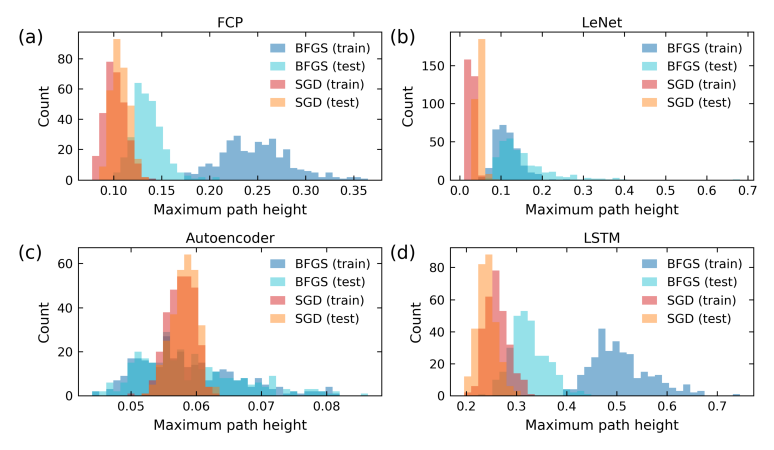}
  \caption{\textbf{Distribution of barrier heights along optimized paths on the training landscape.} Shown are distributions of the \texttt{FourierPathFinder} path heights (Eq.~\eqref{H}) for FCP (a), LeNet (b), Autoencoder (c), and LSTM (d).
  Histograms in each panel show heights of $300$ low-loss paths connecting randomly chosen pairs of optimized parameter vectors in
  $\{ \omega^{\text{train},i}_{\text{BFGS}} \}_{i=1}^{48}$ (navy blue),
  $\{ \omega^{\text{test},i}_{\text{BFGS}} \}_{i=1}^{48}$ (light blue),
  $\{ \omega^{\text{train},i}_{\text{SGD}} \}_{i=1}^{48}$ (light red), and
  $\{ \omega^{\text{test},i}_{\text{SGD}} \}_{i=1}^{48}$ (gold).
  The paths are computed on the training landscape, $\langle l(x^\text{train},\omega) \rangle$.
  }
  \label{Fig:heights:training}
\end{figure}

We find that, with the exception of Autoencoder,
the barrier heights are lower for the SGD solutions compared to BFGS.
This is true for both $\{ \omega^{\text{test},i}_{\text{SGD}} \}_{i=1}^{48}$ and
$\{ \omega^{\text{train},i}_{\text{SGD}} \}_{i=1}^{48}$ on the training landscape (Fig.~\ref{Fig:heights:training}) and $\{ \omega^{\text{test},i}_{\text{SGD}} \}_{i=1}^{48}$ on the
test landscape (Fig.~\ref{Fig:heights:test}). This indicates that SGD solutions are located in smoother, more accessible regions of the loss landscape. In contrast,
BFGS solutions $\{ \omega^{\text{test},i}_{\text{BFGS}} \}_{i=1}^{48}$ and $\{ \omega^{\text{train},i}_{\text{BFGS}} \}_{i=1}^{48}$ are characterized by higher barriers on both landscapes, indicating that those vectors of optimized parameters are more isolated from one another.
Note that Autoencoder is probably an exception because it does not exhibit strong BFGS overfitting prominent in the other three NN architectures (Fig.~\ref{loss:curves:extra}, Table~\ref{loss:table}).

Interestingly, on the training landscape the BFGS barrier heights between vectors of training weights are only higher than the barrier heights between vectors of test weights in two out of four NN architectures, FCP and LSTM (cf. navy blue and light blue histograms in Fig.~\ref{Fig:heights:training}). This is surprising because BFGS training weights are overfit for FCP, LeNet, and LSTM (Table~\ref{loss:table}). In other words, the L-BFGS-GSS optimizer does not necessarily find more isolated minima with additional training, even if overfitting occurs. Although the same observation is true for SGD (cf. light red and gold histograms in Fig.~\ref{Fig:heights:training}), it is less surprising there due to much weaker signatures of overfitting in the case of SGD optimization.





\subsection{Statistics of optimized NN parameters}

In addition to the analysis of the paths connecting pairs of optimized vectors of NN parameters, we consider the statistics of optimized NN weights and biases.
To this end, we compute the means and standard deviations of the components of $W^\text{train,j}_\text{BFGS}$, $W^\text{test,j}_\text{BFGS}$,
$W^\text{train,j}_\text{SGD}$, $W^\text{test,j}_\text{SGD}$ vectors,
where each $W$ vector is constructed by concatenation of the corresponding
$\{ \omega^i \}_{i=1}^{48}$ set of vectors and $j = 1 \dots 4$ labels NN
architectures (Table~\ref{Table:musigma}). We see that
there is a clear difference between SGD and BFGS weights, with the latter
characterized by larger standard deviations $\sigma$. Thus, BFGS fitting weights tend to be more spread out in parameter space.

Another way to see the extent of the spread is to compare the $L_2$ lengths of the SGD and BFGS weight vectors.
We focus first on the comparison between SGD test and BFGS training 
weights since the latter are overfit (except in the Autoencoder; Figs.~\ref{loss:curves} and \ref{loss:curves:extra}, Table~\ref{loss:table}), enabling us to contrast 
SGD weight sets that one would use in practice with low-loss, non-generalizable solutions obtained by L-BFGS-GSS.
Specifically, we define $\Bar{\omega}_{\text{BFGS}} = \frac{1}{48} \sum_{i=1}^{48} \omega^{\text{test},i}_{\text{BFGS}}$ and $\Bar{\omega}_{\text{SGD}} = \frac{1}{48} \sum_{i=1}^{48} \omega^{\text{test},i}_{\text{SGD}}$ as the centroids of the optimized weight vector sets and use $\Bar{\omega}=\frac{1}{2}(\Bar{\omega}_{\text{BFGS}}+\Bar{\omega}_{\text{SGD}})$ as the common origin of all weight vectors.
For each individual weight vector $\omega^i$, we calculate its $L_2$ distance from the origin as $|\omega^i-\Bar{\omega}|$.

\begin{figure}[!htb]
  \centering
  \includegraphics[width=\textwidth]{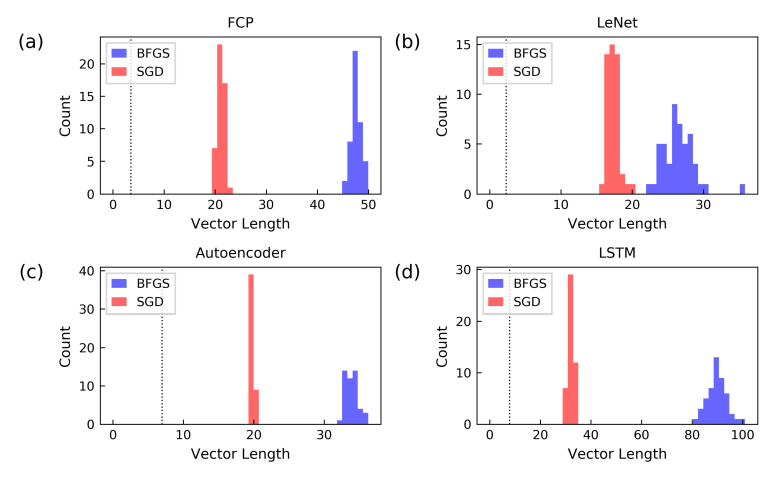}
  \caption{\textbf{Distributions of SGD test and BFGS training weight vector lengths.}
  Shown are the histograms of $L_2$ distances between individual weight vectors
  $\omega_i$ and the common origin $\Bar{\omega}$, $|\omega^i-\Bar{\omega}|$.
  Distributions of the BFGS training ($\{ \omega^{\text{train},i}_{\text{BFGS}} \}_{i=1}^{48}$) and SGD test ($\{ \omega^{\text{test},i}_{\text{SGD}} \}_{i=1}^{48}$) weight vector lengths are plotted in blue and light red, respectively, for FCP (a), LeNet (b), Autoencoder (c), and LSTM (d).
  Dotted vertical lines indicate the positions of $|\bar{\omega}_{\text{BFGS}} - \bar{\omega}| = |\bar{\omega}_{\text{SGD}} - \bar{\omega}|$.
  }
  \label{Fig:L2}
\end{figure}

The histogram of $L_2$ distances shows that, as expected from Table~\ref{Table:musigma},
BFGS weight vectors tend to be longer than SGD weight vectors (Fig.~\ref{Fig:L2}).
This indicates that optimized weight vectors found by L-BFGS-GSS are more widely dispersed compared to the SGD solutions, which form a more compact distribution. The distance between $\Bar{\omega}_{\text{BFGS}}$ and $\Bar{\omega}_{\text{SGD}}$ is small compared to the spread of vector lengths within each group (cf. vertical dotted lines in Fig.~\ref{Fig:L2}), indicating that there is no strongly preferred direction in the parameter space.
Next, we consider SGD and BFGS test weights, as those are the weight vector sets  one would use in practice (Fig.~\ref{Fig:L2:test:test}). We observe that, as might be expected, the SGD and BFGS vector lengths become less different for FCP and LeNet. Surprisingly, the gap between vector length histograms remains nearly the same for Autoencoder and LSTM, despite the latter being overfit in going from test to training weight sets.

\subsection{Low-dimensional projections of optimized weight vectors}

\begin{figure}[!htb]
  \centering
  \includegraphics[width=0.9\textwidth]{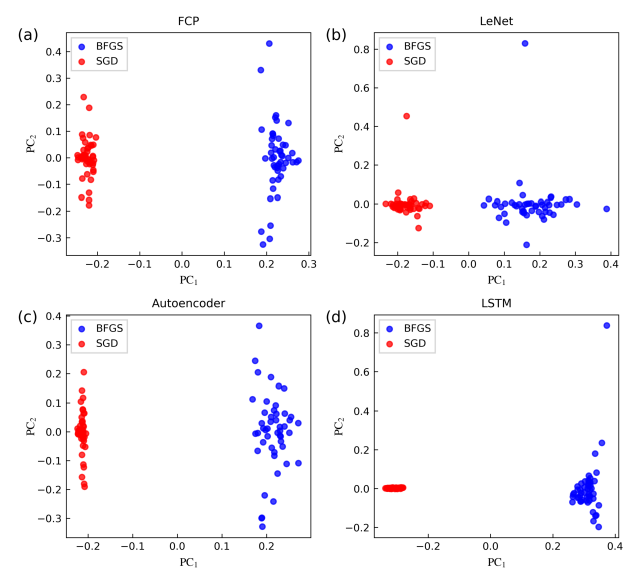}
  \caption{\textbf{kPCA projections of SGD test and BFGS training weight vectors.} Shown are the first two principal components, $\text{PC}_1$ and $\text{PC}_2$, obtained by kPCA with the RBF kernel (Methods). The projections are applied to $\{ \omega^{\text{train},i}_{\text{BFGS}} \}_{i=1}^{48}$ (blue points) and $\{ \omega^{\text{test},i}_{\text{SGD}} \}_{i=1}^{48}$ (red points) sets of optimized weight vectors, for FCP (a), LeNet (b), Autoencoder (c), and LSTM (d).
  }
  \label{Fig:PCA}
\end{figure}

We checked whether the differences between BFGS training and SGD test weight vectors can be detected using principal component analysis (PCA) -- a dimensionality reduction technique often used in data visualization~\cite{Bishop_2006}. Specifically, we have applied kernel PCA (kPCA; see Methods for details) to $\{ \omega^{\text{train},i}_{\text{BFGS}} \}_{i=1}^{48}$ and $\{ \omega^{\text{test},i}_{\text{SGD}} \}_{i=1}^{48}$ sets of optimized weight vectors for each NN architecture (Fig.~\ref{Fig:PCA}). In all four cases, we see clear separation of BFGS and SGD vectors projected onto two first principal components $\text{PC}_1$ and $\text{PC}_2$. Interestingly, the separation is predominantly along the first principal component, indicating that the BFGS and SGD weight vectors form two distinct clusters in the parameter space. Moreover, the separation is only observed with the RBF kernel and disappears when other kernels such as polynomial and sigmoid are used, or when standard PCA is employed (data not shown). Thus, the separation is radial rather than along a preferred direction, consistent with the larger BFGS vector lengths and the absence of preferred directions in Fig.~\ref{Fig:L2}.

When kPCA is applied to BFGS and SGD test weight vectors (Fig.~\ref{Fig:PCA:xtra}), cluster separation nearly disappears for FCP and LeNet
but persists for Autoencoder and LSTM, in agreement with Fig.~\ref{Fig:L2:test:test}.



\section{Discussion and Conclusion}

The results presented here demonstrate that the balance between generalization and overfitting profoundly influences the nature of optimized neural network solutions. Consistent with previous observations~\cite{keskar2017largebatchtrainingdeeplearning,10.5555/3294996.3295170,10.5555/3327345.3327403}, we find that SGD produces more generalizable solutions that occupy flatter, more connected basins of the loss landscape, whereas L-BFGS-GSS solutions consist of sharper, more isolated minima separated by higher barriers. The latter behavior is particularly pronounced if the L-BFGS-GSS optimizer is allowed to overfit, converging to solutions which have much lower loss compared to SGD (Table~\ref{loss:table}).

These observations are confirmed by the analysis of the paths connecting pairs of optimized weight vectors in multi-dimensional parameter space.
We have developed a novel algorithm, \texttt{FourierPathFinder}, which uses a Fourier expansion of paths combined with stochastic gradient optimization to find low-height paths 
between two points on arbitrary loss or energy landscapes. We find that SGD solutions
are typically separated by lower energy barriers than those obtained with L-BFGS-GSS (Figs.~\ref{Fig:heights:training},~\ref{Fig:heights:test}).
The lower barrier heights encountered along SGD paths suggest that SGD solutions tend to lie in broad, smoothly connected valleys of the loss surface, whereas L-BFGS-GSS solutions appear to be embedded in steeper, more difficult-to-navigate regions. Thus, our path analysis supports the view that SGD tends to converge to flatter minima, a hallmark of better generalization, while L-BFGS-GSS, a quasi-Newton method guided by curvature information, is more prone to settling in narrow, high-curvature basins that fit the training data very well but generalize poorly.


This interpretation is reinforced by the analysis of the lengths of  optimized weight vectors (Figs.~\ref{Fig:L2},~\ref{Fig:L2:test:test}; Table~\ref{Table:musigma}). SGD solutions cluster more compactly near a common centroid,
while L-BFGS-GSS solutions are distributed farther from the center, forming a larger-radius shell in parameter space. There appear to be no strongly preferred directions within the SGD and BFGS shells -- the main difference is in the vector lengths, explained at least in part by the larger magnitudes of BFGS vector
components (Table~\ref{Table:musigma}).
Interestingly, the centroids of the SGD and BFGS weight vectors are relatively close to one another, supporting the idea of a nested, shell-like organization of SGD and BFGS weight vectors in the parameter space.

Further evidence of these structural differences emerges from visualization of
multi-dimensional weight vectors based on kPCA projections (Figs.~\ref{Fig:PCA},~\ref{Fig:PCA:xtra}).
kPCA analysis with a spherically symmetric, non-linear RBF kernel reveals clear separation between the solution sets produced by SGD and L-BFGS-GSS, especially when the latter algorithm is allowed to overfit. This observation implies that the two optimizers converge to distinct, nonlinearly separable manifolds in parameter space rather than to nearby points within a single connected region. 


Taken together, our results show that SGD solutions occupy relatively compact, flatter regions of the loss landscape, while L-BFGS-GSS solutions concentrate within a larger-diameter shell corresponding to higher-curvature, less accessible minima. This radial organization supports the idea that stochastic gradient based methods tend to find central, robust basins in parameter space, while deterministic quasi-Newton methods converge toward more spread-out, less generalizable minima. Thus, the choice of the
optimizer affects not only convergence speed but also the qualitative nature of the resulting solutions. The differences between SGD and L-BFGS-GSS reflect
fundamentally different optimization dynamics that guide each method toward distinct regions of the loss landscape.

In practical terms, smoother connectivity between SGD minima facilitates model averaging and transfer compared to the L-BFGS-GSS approach. 
More broadly, the low-barrier connectivity, compact clustering, and nonlinear separability of generalizable solutions provide a geometric foundation for understanding why flatter minima, favored by SGD, tend to yield more robust performance in neural networks.
In summary, our findings underscore how optimizer choice in machine-learning contexts affects not only the efficiency of convergence to low-loss solutions, but also the geometry and diversity of the solutions themselves -- with potential consequences for generalization and robustness.




\section*{Software and Data Availability}

\noindent
The neural network training and loss landscape analysis software was written in Python and is available via GitHub at \texttt{https://github.com/jy856-jpg/path-finding}.

\section*{Acknowledgments}

J.Y. and A.V.M. acknowledge financial and logistical support from the Center for Quantitative Biology, Rutgers University. The authors are grateful to the Office of Advanced Research Computing (OARC) at Rutgers University for providing access to the Amarel cluster. 

\bibliography{sample}

@misc{tieleman2012rmsprop,
  title={{Lecture 6.5 -- RMSProp: Divide the gradient by a running average of its recent magnitude}},
  author={Tieleman, Tijmen and Hinton, Geoffrey},
  year={2012},
  pages={26--31},
  howpublished={COURSERA: Neural Networks for Machine Learning}
}

@article{Feng:2021,
  title={The inverse variance-flatness relation in stochastic gradient descent is critical for finding flat minima},
  author={Yu Feng and Yuhai Tu},
  journal={Proc. Nat. Acad. Sci. USA},
  volume={118},
  pages={e2015617118},
  year={2021}
}

@book{Goodfellow2016,
    title={Deep Learning},
    author={Ian Goodfellow and Yoshua Bengio and Aaron Courville},
    publisher={MIT Press},
    address = {Cambridge, MA, USA},
    year={2016}
}

@book{Crow1970,
        author = {J. F. Crow and M. Kimura},
        title = {An Introduction to Population Genetics Theory},
        year = {1970},
        publisher = {Harper and Row},
        address = {New York, NY, USA}
}

@book{Kimura1983,
        author = "M. Kimura",
        title = "The Neutral Theory of Molecular Evolution",
        publisher = "Cambridge University Press",
        address = "Cambridge, UK",
        year = "1983"
}

@book{Gillespie2004,
        author = "J.H. Gillespie",
        title = "Population Genetics: A Concise Guide",
        publisher = "The Johns Hopkins University Press",
        address = "Baltimore, MD, USA",
        year = "2004",
}

@article{Dill:2008,
  title={The Protein Folding Problem},
  author={Dill, Ken A. and Ozkan, S. Banu and Shell, M. Scott and Weikl, Thomas R.},
  journal={Ann. Rev. Biophys.},
  volume={37},
  pages={289--316},
  year={2008}
}

@article{Onuchic:2004,
  title = {Theory of protein folding},
  author = {J. N. Onuchic and P. G. Wolynes},
  journal = {Curr. Op. Struct. Biol.},
  volume = {14},
  number = {},
  pages = {70--75},
  year = {2004}
}

@article{Scholkopf1998,
  author={Scholkopf, Bernhard and Smola, Alexander and Muller, Klaus-Robert},
  journal={Neural Computation}, 
  title={Nonlinear Component Analysis as a Kernel Eigenvalue Problem}, 
  year={1998},
  volume={10},
  number={5},
  pages={1299--1319}
}

@misc{kingma2014adam,
  title={Adam: A Method for Stochastic Optimization},
  author={Kingma, Diederik P. and Ba, Jimmy Lei},
  year={2014},
  eprint={1412.6980},
  archivePrefix={arXiv},
  primaryClass={cs.LG},
  url={https://arxiv.org/abs/1412.6980}
}

@article{Mehta:2018dln,
    author = "Mehta, Pankaj and Bukov, Marin and Wang, Ching-Hao and Day, Alexandre G. R. and Richardson, Clint and Fisher, Charles K. and Schwab, David J.",
    title = {A high-bias, low-variance introduction to {M}achine {L}earning for physicists},
    journal = "Phys. Rep.",
    volume = "810",
    pages = "1--124",
    year = "2019"
}

@article{Broyden1970,
  author = {Broyden, C. G.},
  title = {Quasi-{N}ewton Methods and Their Application to Function Minimisation},
  journal = {Mathematics of Computation},
  volume = {24},
  pages = {365--381},
  year = {1970}
}

@article{Fletcher1970,
  author = {Fletcher, R.},
  title = {A New Approach to Variable Metric Algorithms},
  journal = {The Computer Journal},
  volume = {13},
  pages = {317--322},
  year = {1970}
}

@article{Goldfarb1970,
  author = {Goldfarb, D.},
  title = {A Family of Variable-Metric Methods Derived by Variational Means},
  journal = {Mathematics of Computation},
  volume = {24},
  pages = {23--26},
  year = {1970}
}

@article{Shanno1970,
  author = {Shanno, D. F.},
  title = {Conditioning of Quasi-{N}ewton Methods for Function Minimization},
  journal = {Mathematics of Computation},
  volume = {24},
  pages = {647--656},
  year = {1970}
}

@misc{ruder2017,
      title={An overview of gradient descent optimization algorithms}, 
      author={Sebastian Ruder},
      year={2017},
      eprint={1609.04747},
      archivePrefix={arXiv},
      primaryClass={cs.LG},
      url={https://arxiv.org/abs/1609.04747}
}

@article{10.1162/neco.1997.9.8.1735,
    author = {Hochreiter, Sepp and Schmidhuber, Jurgen},
    title = {Long Short-Term Memory},
    journal = {Neural Computation},
    volume = {9},
    number = {8},
    pages = {1735--1780},
    year = {1997},
    doi = {10.1162/neco.1997.9.8.1735}
}

@misc{le2015simplewayinitializerecurrent,
      title={A Simple Way to Initialize Recurrent Networks of Rectified Linear Units},
      author={Quoc V. Le and Navdeep Jaitly and Geoffrey E. Hinton},
      year={2015},
      eprint={1504.00941},
      archivePrefix={arXiv},
      primaryClass={cs.NE},
      url={https://arxiv.org/abs/1504.00941}, 
}

@inbook{10.5555/104279.104293,
author = {Rumelhart, D. E. and Hinton, G. E. and Williams, R. J.},
title = {Learning internal representations by error propagation},
year = {1986},
isbn = {026268053X},
publisher = {MIT Press},
address = {Cambridge, MA, USA},
booktitle = {Parallel Distributed Processing: Explorations in the Microstructure of Cognition, Vol. 1: Foundations},
pages = {318--362}
}

@article{10.1145/3446776,
author = {Zhang, Chiyuan and Bengio, Samy and Hardt, Moritz and Recht, Benjamin and Vinyals, Oriol},
title = {Understanding deep learning (still) requires rethinking generalization},
year = {2021},
issue_date = {March 2021},
publisher = {Association for Computing Machinery},
address = {New York, NY, USA},
volume = {64},
doi = {10.1145/3446776},
journal = {Commun. ACM},
pages = {107--115}
}

@InProceedings{pmlr-v80-draxler18a,
  title = 	 {Essentially No Barriers in Neural Network Energy Landscape},
  author =       {Draxler, Felix and Veschgini, Kambis and Salmhofer, Manfred and Hamprecht, Fred},
  booktitle = 	 {Proceedings of the 35th International Conference on Machine Learning},
  pages = 	 {1309--1318},
  year = 	 {2018},
  editor = 	 {Dy, Jennifer and Krause, Andreas},
  volume = 	 {80},
  series = 	 {Proceedings of Machine Learning Research},
  publisher =    {PMLR},
  url = 	 {https://proceedings.mlr.press/v80/draxler18a.html}
}

@inproceedings{10.5555/3327546.3327556,
author = {Garipov, Timur and Izmailov, Pavel and Podoprikhin, Dmitrii and Vetrov, Dmitry and Wilson, Andrew Gordon},
title = {Loss surfaces, mode connectivity, and fast ensembling of {DNNs}},
year = {2018},
publisher = {Curran Associates Inc.},
address = {Red Hook, NY, USA},
booktitle = {Proceedings of the 32nd International Conference on Neural Information Processing Systems},
pages = {8803--8812},
location = {Montreal, Canada},
series = {NIPS'18}
}

@book{Bishop_2006,
title={Pattern recognition and machine learning},
publisher={Springer Science+Business Media, LLC, New York, NY, USA},
author={Bishop, Christopher M.},
year={2006}
}

@book{Nocedal_Wright_2006,
place={New York, NY},
title={Numerical optimization},
publisher={Springer Science+Business Media, LLC, New York, NY, USA}, author={Nocedal, Jorge and Wright, Stephen J.},
year={2006}
}

@misc{keskar2017largebatchtrainingdeeplearning,
      title={On Large-Batch Training for Deep Learning: Generalization Gap and Sharp Minima}, 
      author={Nitish Shirish Keskar and Dheevatsa Mudigere and Jorge Nocedal and Mikhail Smelyanskiy and Ping Tak Peter Tang},
      year={2017},
      eprint={1609.04836},
      archivePrefix={arXiv},
      primaryClass={cs.LG},
      url={https://arxiv.org/abs/1609.04836}, 
}

@inproceedings{NIPS1988_a97da629,
 author = {Denker, John and Gardner, W. and Graf, Hans and Henderson, Donnie and Howard, R. and Hubbard, W. and Jackel, L. D. and Baird, Henry and Guyon, Isabelle},
 booktitle = {{Advances in Neural Information Processing Systems}},
 editor = {D. Touretzky},
 pages = {},
 publisher = {Morgan-Kaufmann},
 title = {Neural Network Recognizer for Hand-Written Zip Code Digits},
 url = {https://proceedings.neurips.cc/paper_files/paper/1988/file/a97da629b098b75c294dffdc3e463904-Paper.pdf},
 volume = {1},
 year = {1988}
}

@ARTICLE{726791,
  author={LeCun, Y. and Bottou, L. and Bengio, Y. and Haffner, P.},
  journal={Proceedings of the IEEE}, 
  title={Gradient-based learning applied to document recognition}, 
  year={1998},
  volume={86},
  number={11},
  pages={2278-2324},
  doi={10.1109/5.726791}}

@ARTICLE{6296535,
  author={Deng, Li},
  journal={IEEE Signal Processing Magazine}, 
  title={The {MNIST} Database of Handwritten Digit Images for Machine Learning Research [Best of the {W}eb]}, 
  year={2012},
  volume={29},
  number={6},
  pages={141--142},
  doi={10.1109/MSP.2012.2211477}}

@misc{wei2019noiseaffectshessianspectrum,
      title={How noise affects the Hessian spectrum in overparameterized neural networks}, 
      author={Mingwei Wei and David J Schwab},
      year={2019},
      eprint={1910.00195},
      archivePrefix={arXiv},
      primaryClass={cs.LG},
      url={https://arxiv.org/abs/1910.00195}, 
}

@article{doi:10.1137/16M1080173,
author = {Bottou, L\'{e}on and Curtis, Frank E. and Nocedal, Jorge},
title = {Optimization Methods for Large-Scale Machine Learning},
journal = {SIAM Review},
volume = {60},
number = {2},
pages = {223--311},
year = {2018},
doi = {10.1137/16M1080173}
}

@article{doi:10.1137/1011036,
author = {Wolfe, Philip},
title = {Convergence Conditions for Ascent Methods},
journal = {SIAM Review},
volume = {11},
number = {2},
pages = {226--235},
year = {1969},
doi = {10.1137/1011036}
}

@book{press2007numerical,
  title={{Numerical Recipes: The Art of Scientific Computing}},
  author={Press, William H and Teukolsky, Saul A and Vetterling, William T and Flannery, Brian P},
  year={2007},
  edition={3rd},
  publisher={Cambridge University Press, Cambridge, UK}
}

@article{Chaudhari_2019,
doi = {10.1088/1742-5468/ab39d9},
url = {https://dx.doi.org/10.1088/1742-5468/ab39d9},
year = {2019},
publisher = {IOP Publishing and SISSA},
volume = {2019},
pages = {124018},
author = {Chaudhari, Pratik and Choromanska, Anna and Soatto, Stefano and LeCun, Yann and Baldassi, Carlo and Borgs, Christian and Chayes, Jennifer and Sagun, Levent and Zecchina, Riccardo},
title = {{Entropy-SGD: biasing gradient descent into wide valleys}},
journal = {Journal of Statistical Mechanics: Theory and Experiment},
}

@misc{jastrzębski2018factorsinfluencingminimasgd,
      title={Three Factors Influencing Minima in {SGD}}, 
      author={Stanislaw Jastrzebski and Zachary Kenton and Devansh Arpit and Nicolas Ballas and Asja Fischer and Yoshua Bengio and Amos Storkey},
      year={2018},
      eprint={1711.04623},
      archivePrefix={arXiv},
      primaryClass={cs.LG},
      url={https://arxiv.org/abs/1711.04623}, 
}

@article{Robbins_Monro_1951,
title={A Stochastic Approximation Method},
volume={22},
doi={10.1214/aoms/1177729586},
journal={The Annals of Mathematical Statistics},
author={Robbins, Herbert and Monro, Sutton},
year={1951},
pages={400--407}
}

@misc{goodfellow2015qualitativelycharacterizingneuralnetwork,
      title={Qualitatively characterizing neural network optimization problems}, 
      author={Ian J. Goodfellow and Oriol Vinyals and Andrew M. Saxe},
      year={2015},
      eprint={1412.6544},
      archivePrefix={arXiv},
      primaryClass={cs.NE},
      url={https://arxiv.org/abs/1412.6544}, 
}

@inproceedings{NEURIPS2018_a41b3bb3,
 author = {Li, Hao and Xu, Zheng and Taylor, Gavin and Studer, Christoph and Goldstein, Tom},
 booktitle = {{Advances in Neural Information Processing Systems}},
 editor = {S. Bengio and H. Wallach and H. Larochelle and K. Grauman and N. Cesa-Bianchi and R. Garnett},
 pages = {},
 publisher = {Curran Associates, Inc.},
 title = {Visualizing the Loss Landscape of Neural Nets},
 volume = {31},
 year = {2018}
}

@InProceedings{pmlr-v97-ghorbani19b,
  title = 	 {An Investigation into Neural Net Optimization via {H}essian Eigenvalue Density},
  author =       {Ghorbani, Behrooz and Krishnan, Shankar and Xiao, Ying},
  booktitle = 	 {Proceedings of the 36th International Conference on Machine Learning},
  pages = 	 {2232--2241},
  year = 	 {2019},
  editor = 	 {Chaudhuri, Kamalika and Salakhutdinov, Ruslan},
  volume = 	 {97},
  series = 	 {Proceedings of Machine Learning Research},
  publisher =    {PMLR},
}

@inproceedings{10.5555/3294996.3295170,
author = {Wilson, Ashia C. and Roelofs, Rebecca and Stern, Mitchell and Srebro, Nathan and Recht, Benjamin},
title = {The marginal value of adaptive gradient methods in machine learning},
year = {2017},
publisher = {Curran Associates Inc.},
address = {Red Hook, NY, USA},
booktitle = {Proceedings of the 31st International Conference on Neural Information Processing Systems},
pages = {4151--4161},
location = {Long Beach, California, USA},
series = {NIPS'17}
}

@inproceedings{10.5555/3327345.3327403,
author = {Yao, Zhewei and Gholami, Amir and Keutzer, Kurt and Mahoney, Michael W.},
title = {Hessian-based analysis of large batch training and robustness to adversaries},
year = {2018},
publisher = {Curran Associates Inc.},
address = {Red Hook, NY, USA},
booktitle = {Proceedings of the 32nd International Conference on Neural Information Processing Systems},
pages = {4954--4964},
location = {Montreal, Canada},
series = {NIPS'18}
}

@misc{sonthalia2024deepneuralnetworksolutions,
      title={Do Deep Neural Network Solutions Form a Star Domain?}, 
      author={Ankit Sonthalia and Alexander Rubinstein and Ehsan Abbasnejad and Seong Joon Oh},
      year={2024},
      eprint={2403.07968},
      archivePrefix={arXiv},
      primaryClass={cs.LG},
      url={https://arxiv.org/abs/2403.07968}, 
}

@misc{zhou2019sgdconvergesglobalminimum,
      title={{SGD} Converges to Global Minimum in Deep Learning via Star-convex Path}, 
      author={Yi Zhou and Junjie Yang and Huishuai Zhang and Yingbin Liang and Vahid Tarokh},
      year={2019},
      eprint={1901.00451},
      archivePrefix={arXiv},
      primaryClass={cs.LG},
      url={https://arxiv.org/abs/1901.00451}, 
}

\newpage
\section*{Supplementary Materials} \label{sec:sup_fig}

\renewcommand{\thefigure}{S\arabic{figure}}
\setcounter{figure}{0}

\renewcommand{\thetable}{S\arabic{table}}
\setcounter{table}{0}

\begin{figure}[!htb]
  \centering
  \includegraphics[width=0.95\textwidth]{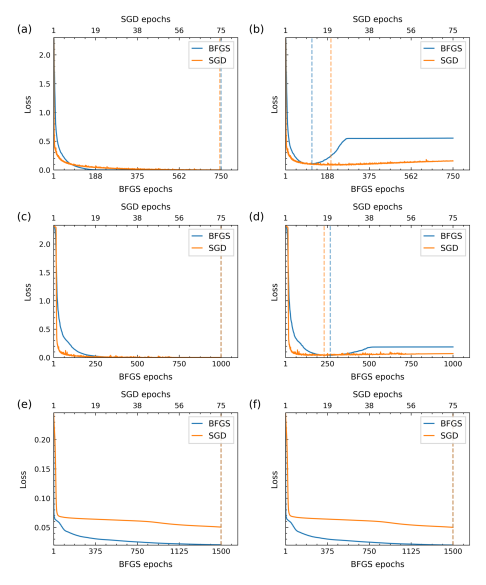}
  \caption{\textbf{Additional examples of loss curves.}
  Same as Fig.~\ref{loss:curves} but for FCP (training: a, test: b), LeNet (training: c, test: d), and Autoencoder (training: e, test: f) NN architectures.
  }
  \label{loss:curves:extra}
\end{figure}

\begin{figure}[!htb]
  \centering
  \includegraphics[width=\textwidth]{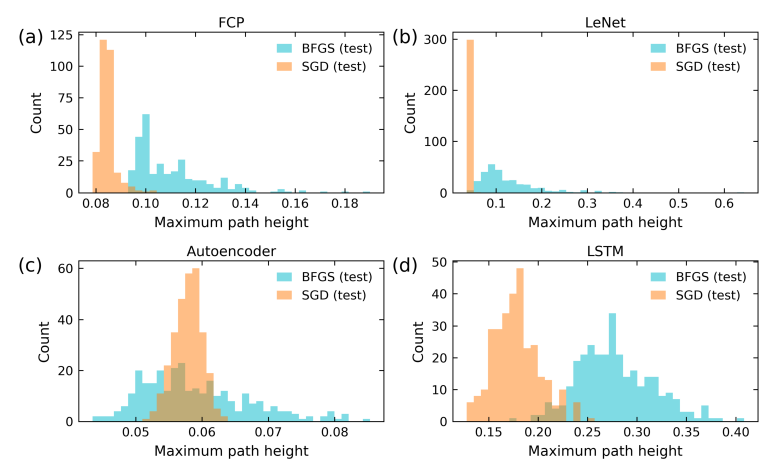}
  \caption{\textbf{Distribution of barrier heights along optimized paths on the test landscape.} Shown are distributions of the \texttt{FourierPathFinder} path heights (Eq.~\eqref{H}) for FCP (a), LeNet (b), Autoencoder (c), and LSTM (d).
  Histograms in each panel show heights of $300$ low-loss paths connecting randomly chosen pairs of optimized parameter vectors in
  $\{ \omega^{\text{test},i}_{\text{BFGS}} \}_{i=1}^{48}$ (light blue) and
  $\{ \omega^{\text{test},i}_{\text{SGD}} \}_{i=1}^{48}$ (gold).
  The paths are computed on the test landscape, $\langle l(x^\text{test},\omega) \rangle$.
  }
  \label{Fig:heights:test}
\end{figure}

\begin{figure}[!htb]
  \centering
  \includegraphics[width=\textwidth]{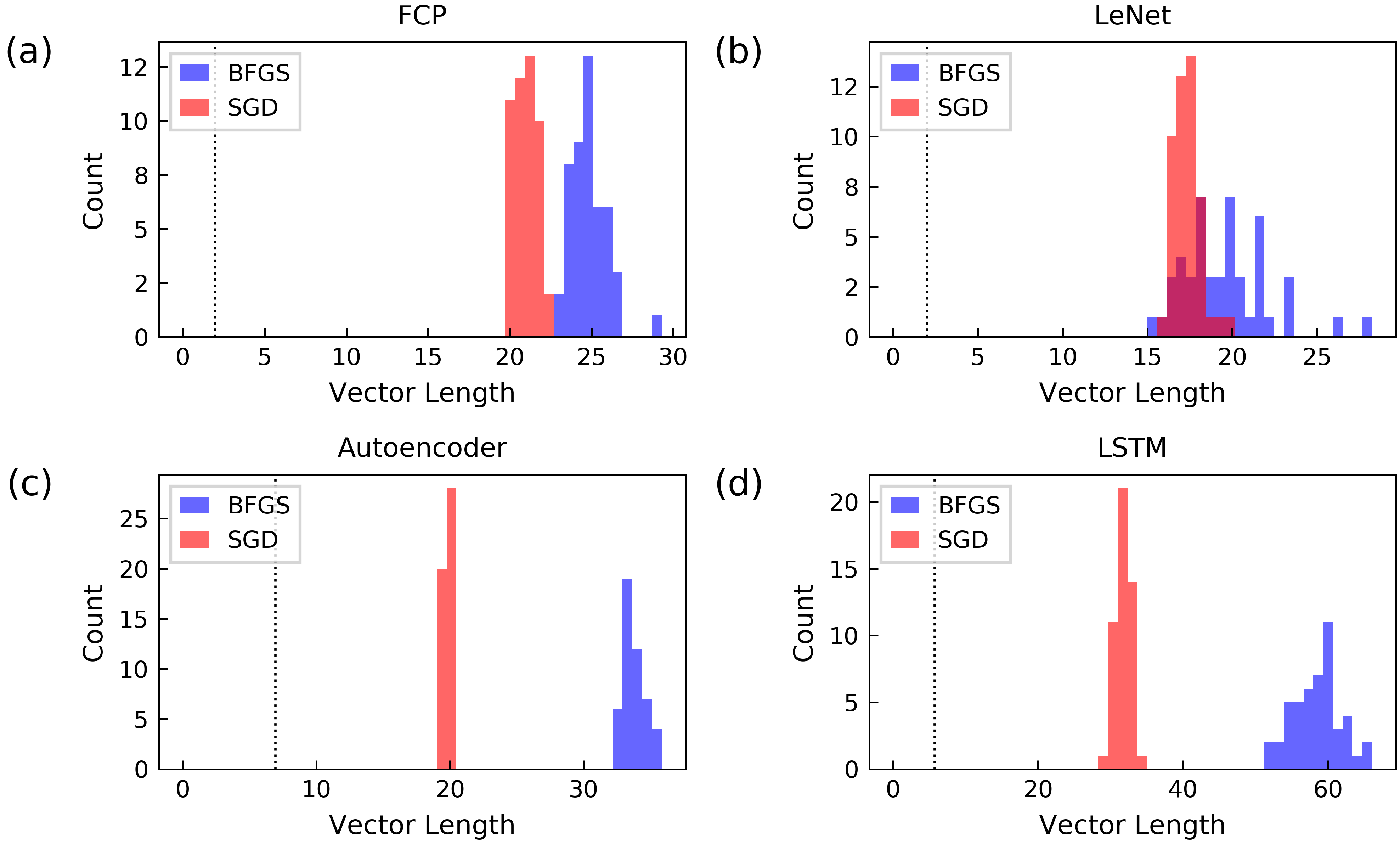}
  \caption{\textbf{Distributions of SGD and BFGS test weight vector lengths.}
  Shown are the histograms of $L_2$ distances between individual weight vectors
  $\omega_i$ and the common origin $\Bar{\omega}$, $|\omega^i-\Bar{\omega}|$.
  Distributions of the BFGS ($\{ \omega^{\text{test},i}_{\text{BFGS}} \}_{i=1}^{48}$) and SGD ($\{ \omega^{\text{test},i}_{\text{SGD}} \}_{i=1}^{48}$) test weight vector lengths are plotted in blue and light red, respectively, for FCP (a), LeNet (b), Autoencoder (c), and LSTM (d).
  Dotted vertical lines indicate the positions of $|\bar{\omega}_{\text{BFGS}} - \bar{\omega}| = |\bar{\omega}_{\text{SGD}} - \bar{\omega}|$.
  }
  \label{Fig:L2:test:test}
\end{figure}

\begin{figure}[!htb]
  \centering
  \includegraphics[width=0.9\textwidth]{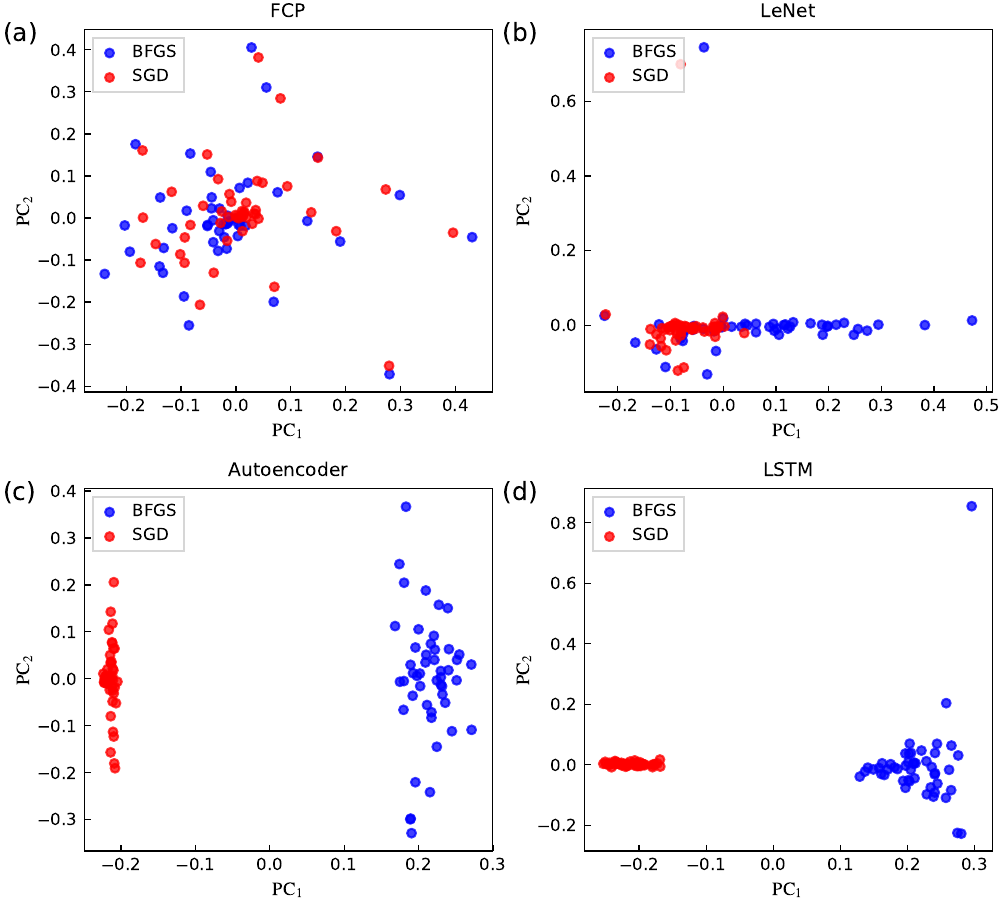}
  \caption{\textbf{kPCA projections of SGD and BFGS test weight vectors.} Shown are the first two principal components, $\text{PC}_1$ and $\text{PC}_2$, obtained by kPCA with the RBF kernel (Methods). The projections are applied to $\{ \omega^{\text{test},i}_{\text{BFGS}} \}_{i=1}^{48}$ (blue points) and $\{ \omega^{\text{test},i}_{\text{SGD}} \}_{i=1}^{48}$ (red points) sets of optimized weight vectors, for FCP (a), LeNet (b), Autoencoder (c), and LSTM (d).
  }
  \label{Fig:PCA:xtra}
\end{figure}


\clearpage
\newpage

\begin{table}[h]
\centering
\caption{\textbf{Statistics of training and test weight distributions.}
Means ($\mu$) and standard deviations ($\sigma$) of
$W^\text{train}_\text{SGD}$, $W^\text{train}_\text{BFGS}$, $W^\text{test}_\text{SGD}$, $W^\text{test}_\text{BFGS}$ -- combined
vectors of optimized weights and biases for each NN architecture.
}
\begin{tabular}{lrrrrrrrr}
\hline
 & \multicolumn{4}{c}{\textbf{Training weights}} & \multicolumn{4}{c}{\textbf{Test weights}} \\
\cline{2-5}\cline{6-9}
NN 
& \multicolumn{2}{c}{SGD} & \multicolumn{2}{c}{L-BFGS-GSS}
& \multicolumn{2}{c}{SGD} & \multicolumn{2}{c}{L-BFGS-GSS} \\
& \multicolumn{1}{c}{$\mu$} & \multicolumn{1}{c}{$\sigma$} & \multicolumn{1}{c}{$\mu$} & \multicolumn{1}{c}{$\sigma$}
& \multicolumn{1}{c}{$\mu$} & \multicolumn{1}{c}{$\sigma$} & \multicolumn{1}{c}{$\mu$} & \multicolumn{1}{c}{$\sigma$} \\
\hline
FCP  
& -1.44e{-03} & 1.39e{-01} & -2.47e{-03} & 2.34e{-01}
& -3.17e{-03} & 1.03e{-01} & -4.11e{-03} & 1.22e{-01} \\

LeNet   
& -3.57e{-03} & 8.77e{-02} & 1.04e{-03}  & 1.08e{-01}
& -2.83e{-03} & 7.06e{-02} & 2.33e{-04} & 8.00e{-02} \\

Autoencoder   
& -7.06e{-03} & 8.54e{-02} & -3.65e{-02} & 1.60e{-01}
& -7.07e{-03} & 8.54e{-02} & -3.65e{-02} & 1.61e{-01} \\

LSTM 
& 1.35e{-02}  & 3.24e{-01} & 1.06e{-02}  & 7.33e{-01}
& 1.16e{-02}  & 2.59e{-01} & 2.92e{-03}  & 4.76e{-01} \\
\hline
\end{tabular}
\label{Table:musigma}
\end{table}

\end{document}